\documentclass[sigconf]{acmart}
\renewcommand\footnotetextcopyrightpermission[1]{} 
\usepackage{multirow}
\usepackage{graphicx}
\usepackage{caption}
\usepackage{fancyhdr}
\AtBeginDocument{%
  \providecommand\BibTeX{{%
    \normalfont B\kern-0.5em{\scshape i\kern-0.25em b}\kern-0.8em\TeX}}}

\setcopyright{acmcopyright}

\begin{document}

\title{Top1 Solution of QQ Browser 2021 Ai Algorithm \\ 
Competition Track 1 : Multimodal Video Similarity }


\author{Zhuoran Ma}
\affiliation{%
  \institution{Baidu. Inc}
  \city{Beijing}
  \country{China}
  }
\email{mzrzju@gmail.com}

\author{Majing Lou}
\affiliation{%
  \institution{Mininglamp. Inc}
  \city{Shanghai}
  \country{China}
  }
\email{loumajing@gmail.com}

\author{Xuan Ouyang}
\affiliation{%
  \institution{Sogou. Inc}
  \city{Beijing}
  \country{China}
  }
\email{oyxuan67373@gmail.com }


\begin{abstract}
  In this paper, we describe the solution to the QQ Browser 2021 Ai Algorithm Competition (AIAC) Track 1. We use the multi-modal transformer model for the video embedding extraction. 
  In the pretrain phase, we train the model with three tasks, (1) Video Tag Classification (VTC), (2) Mask Language Modeling (MLM) and (3) Mask Frame Modeling (MFM).
  In the finetune phase, we train the model with video similarity based on rank normalized human labels.
  Our full pipeline, after ensembling several models, scores 0.852 on the leaderboard, which we achieved the 1st place in the competition. The source codes have been released at Github\footnote{https://github.com/zr2021/2021\_QQ\_AIAC\_Tack1\_1st}.
               
\end{abstract}



\maketitle
\pagestyle{plain}

\section{Introduction}
The 2021 AIAC competition\cite{ref1} is held by QQ browser with support from CCF TCMT and ACM conference CIKM. Track1 is the  Multimodal Video Similarity task. In order to explore video content embedding, the host provides millions of labeled videos in real business and tens of thousands of semantic video similarity annotations. We are challenged to extract the video embedding, and the semantic similarities between videos is defined as cosine similarities.

\subsection{Dataset}
We are provided three kinds of video datasets, (1) pretraining dataset, (2) pairwise dataset and (3) test dataset. In the following, we will describe the these these datasets:

\begin{itemize}
    \item The pretraining dataset contains 1 million video samples.

    \item The pairwise dataset contains 67,899 video pairs. Each pair has a similarity score annotated by human.

    \item The test dataset contains two sets, including test\_a for the preliminary contest and test\_b for the final. These two sets contain 31,514 and 43,027 videos.
\end{itemize}

Each video sample contains several fields. In the following, we  will describe each field:
\begin{itemize}
    \item Title : video title in Chinese.
    \item Frame features : video frame features are extracted from the video at 1 fps, and at most 32 frames. The frame embedding is extracted using ImageNet pretrained EfficientNetB3\cite{ref9}, each frame feature is a 1536 dimension vector.
    \item Video tag id and category id : These ids are manually annotated by human. Notice test dataset doesn't contain tag id and category id.
\end{itemize}

In our pipeline, we only use the video title and the video frame features as our model's input.

\subsection{Evaluation metric}
In the competition, we need to submit the video embedding of the test dataset, the embedding dimension is at most 256.
The cosine similarity of each video pair is calculated based on submitted embedding as video similarity.
The evaluation metric is Spearman's rank correlation, which is calculated between video similarity and human-labeled similarity. To compute Spearman's rank correlation, the first step is convert video similarity $ X_{i} $ and human labels $ Y_{i} $ to rank $x_{i}$, $y_{i}$ respectively. Let $\bar{x}$ is the mean of rank $x$, $\bar{y}$ is the mean of rank $y$, then the Spearman $ \rho $ can be computed as:

$$ \rho = \frac{\sum_{i=1}(x_{i} - \bar{x})(y_{i} - \bar{y})}{\sqrt{\sum_{i=1}(x_{i} - \bar{x})^{2}(y_{i} - \bar{y})^{2}}} $$

\begin{figure*}[htpb]
\centering
\includegraphics[width=15cm]{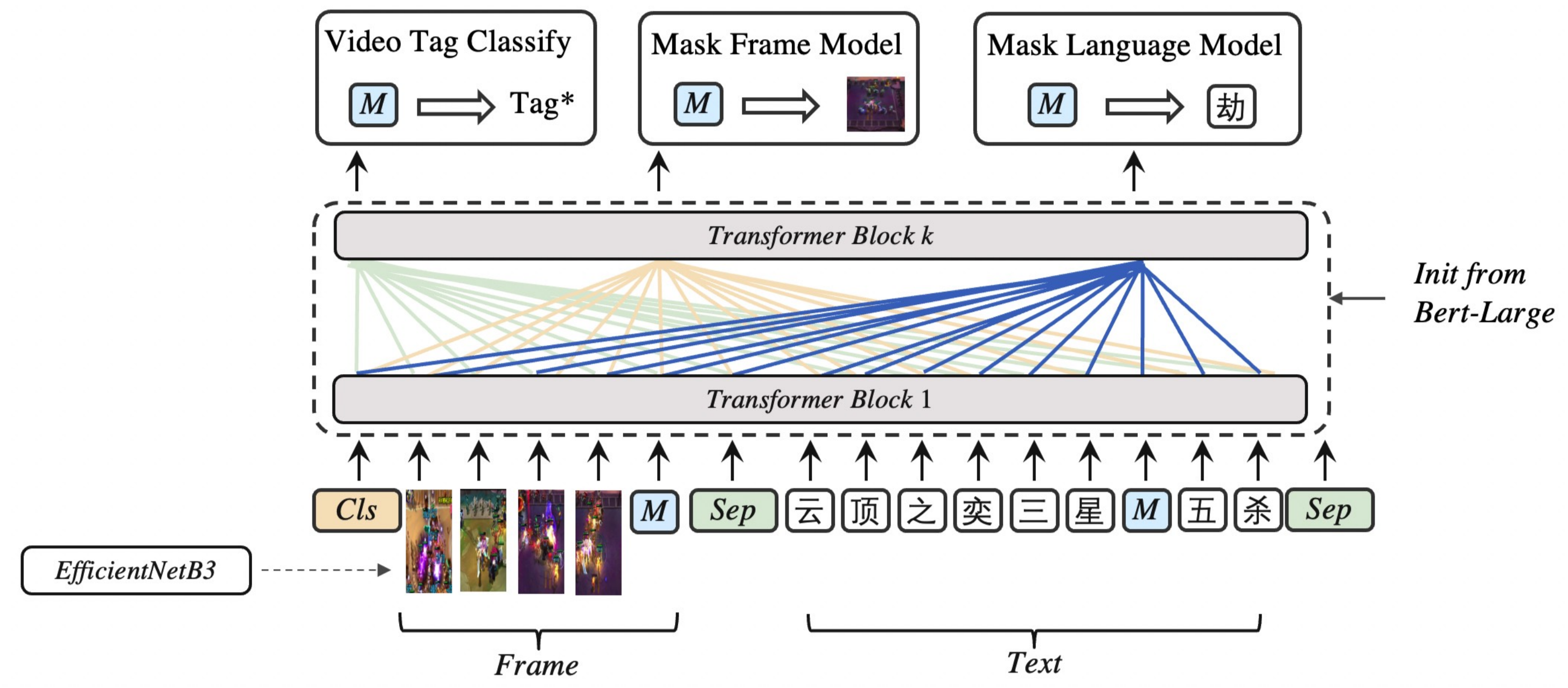}
\caption{Model structure for multi-modal pretrain.}
\label{fig1}
\vskip -0.1in
\end{figure*}


\section{Methodology}
               
\subsection{Modeling}
Inspired from \cite{ref2, ref3, ref4, ref10}, we embed videos into a 256-dimensional feature space as extracted from the pooling layer of Bert models.
The model architecture is illustrated in Figure\ref{fig1}. 
Given a video sample, the model takes the video frame embedding and textual tokens of the video title as the input.
We concatenate special token $[CLS]$, video frame embedding, special token $[SEP]$ and video embedding, then feed them into Bert model to learn multi-modal representation. 
For the pretrain phase, we add a fully connected (fc) layer for each pretrain task.
For the finetune phase, we add a mean pooling layer to extract video embedding and a fc layer to reduce the output's dimension.
                         
\subsection{Pretrain}

\subsubsection{Pretrain task}

The pretrain phase consists of 3 tasks : (1) Video Tag Classify, (2) Mask Language Modeling and (3) Mask Frame Modeling. We compare the downstream performances on the valid set with different pretrain tasks in table 1.

\paragraph{Video tag classify} 
Video tag classify is a multi-label classify task, the goal is to predict which tag does the video belongs to. 
Video tag labels are human-labeled and available in the pretraining dataset and the pairwise dataset which is described in section 1.1.
We use the 10,000 most frequent tags as the ground truths. The final hidden states corresponding to $[CLS]$ token are fed into a classify header to predict tags. 
The loss is the binary cross entropy.

\paragraph{Mask language model}
In the MLM task, we random mask the tokens in video titles and replaced the masked ones with special tokens $[MASK]$.
The goal is to restore these masked tokens based on the video frame and the video titles which the model can attend to. 
For the task setting, we followed the traditional BERT\cite{ref7}, which is 15$\%$ tokens are chosen at random. 
80$\%$ of the time, we replaced the masked input tokens with special tokens $[MASK]$.
10$\%$ of the time, we replaced the masked input tokens with random words.
10$\%$ of the time, we keep the masked input tokens unchanged.
The final hidden states corresponding to the masked token are fed into a classify header and the loss is cross entropy.
               
\paragraph{Mask frame model}
Inspired by \cite{ref3, ref5, ref6}, we add a mask frame modeling (MFM) task to train our model. 
Similar to the MLM task, we random mask a frame by setting this frame feature with an all zeros vector. 
The goal is to restore the frame features based on the model's inputs.
Unlike the MLM task which has a fixed vocabulary, the frame features are continuous and hard to reconstruct.
We use the contrastive learning method to maximize the mutual information between the masked frame features and the original features. 
For the task setting, 15$\%$ frames are chosen at random. 
90$\%$ of the time, we replaced masked input frame features with an all zeros vector.
10$\%$ of the time, we keep the masked input frame features unchanged.
The final hidden states corresponding to the masked frame are fed into a fc layer to extract the frame embedding.
The loss is the noisy contrastive estimation (NCE), and we take the other frames in the same batch as negative.

\paragraph{Multi task pretrain}
We train our model with these three tasks described above simultaneously. The total loss of pretraining is a weighted sum of each task loss, computed as:
$$ L_{total} = L_{vtc} + L_{mlm} + L_{mfm} $$
Notice the video tag classify task has a small gradient magnitude, so it is multiplied by a larger weight.
               
\subsubsection{Pretrain details}
We pretrain three tasks on the pretraining dataset and the pairwise dataset for 40 epochs with 128 batch size.
The learning rate for Bert backbone is 5e-5, and the learning rate for each task header is 5e-4. The scheduler is warm up with cosine decay, warm up ratio is 0.06.

\paragraph{Initialization}
Bert model is initialized from Chinese-BERT-wwm Large\cite{ref8}, which was pretrained from Chinese corpus with whole-word-masking language model task.

\paragraph{Pretrain longer} 
We find it beneficial to pretrain longer and we compare the downstream task performances of different pretrain epochs in table 2. It shows that  longer pretrain has a better downstream task preformance.
         
\paragraph{Model size}     
We compare Bert-base and Bert-large downstream performances in table 3. We can observe that the larger model has a better downstream task performance.

\renewcommand\arraystretch{}
\begin{table}[!h]
\centering

\scalebox{1}{
\begin{tabular}{c | c | c}
\toprule

Pretrain task & Valid Spearman & Valid MSE \\
\midrule
VTC & 0.8786 ± 0.0020 & 0.0308 ± 0.0008\\
VTC + MLM & 0.8812 ± 0.0033 & 0.0300 ± 0.0012  \\
VTC + MLM + MFM & \textbf{0.8858 ± 0.0009} & \textbf{0.0288 ± 0.0005} \\
\bottomrule
\end{tabular}}
\caption{Downstream valid performances of different pretrain task}
\label{table2}
\end{table}
\vspace{-0.7cm}

\renewcommand\arraystretch{}
\begin{table}[!h]
\centering

\scalebox{1}{
\begin{tabular}{c | c | c}
\toprule

Pretrain epochs & Valid Spearman & Valid MSE \\
\midrule
10 & 0.8824 ± 0.0025 & 0.0297 ± 0.0009 \\
20 & 0.8846 ± 0.0008 & 0.0292 ± 0.0005  \\
40 & \textbf{0.8858 ± 0.0009} & \textbf{0.0288 ± 0.0005} \\
\bottomrule
\end{tabular}}
\caption{Downstream valid performances of different pretrain epochs}
\label{table3}
\end{table}
\vspace{-0.7cm}

\renewcommand\arraystretch{}
\begin{table}[!h]
\centering

\scalebox{1}{
\begin{tabular}{c | c |  c | c}
\toprule

Pretrain model & Layers &  Valid Spearman & Valid MSE \\
\midrule
Bert-base & 12 & 0.8791 ± 0.0020 & 0.0306 ± 0.0010 \\
Bert-large & 24 & \textbf{0.8858 ± 0.0009} & \textbf{0.0288 ± 0.0005} \\
\bottomrule
\end{tabular}}
\caption{Downstream valid performances of different model size}
\label{table4}
\end{table}
\vspace{0.2cm}

\begin{figure*}[htpb]
\centering
\includegraphics[width=13cm]{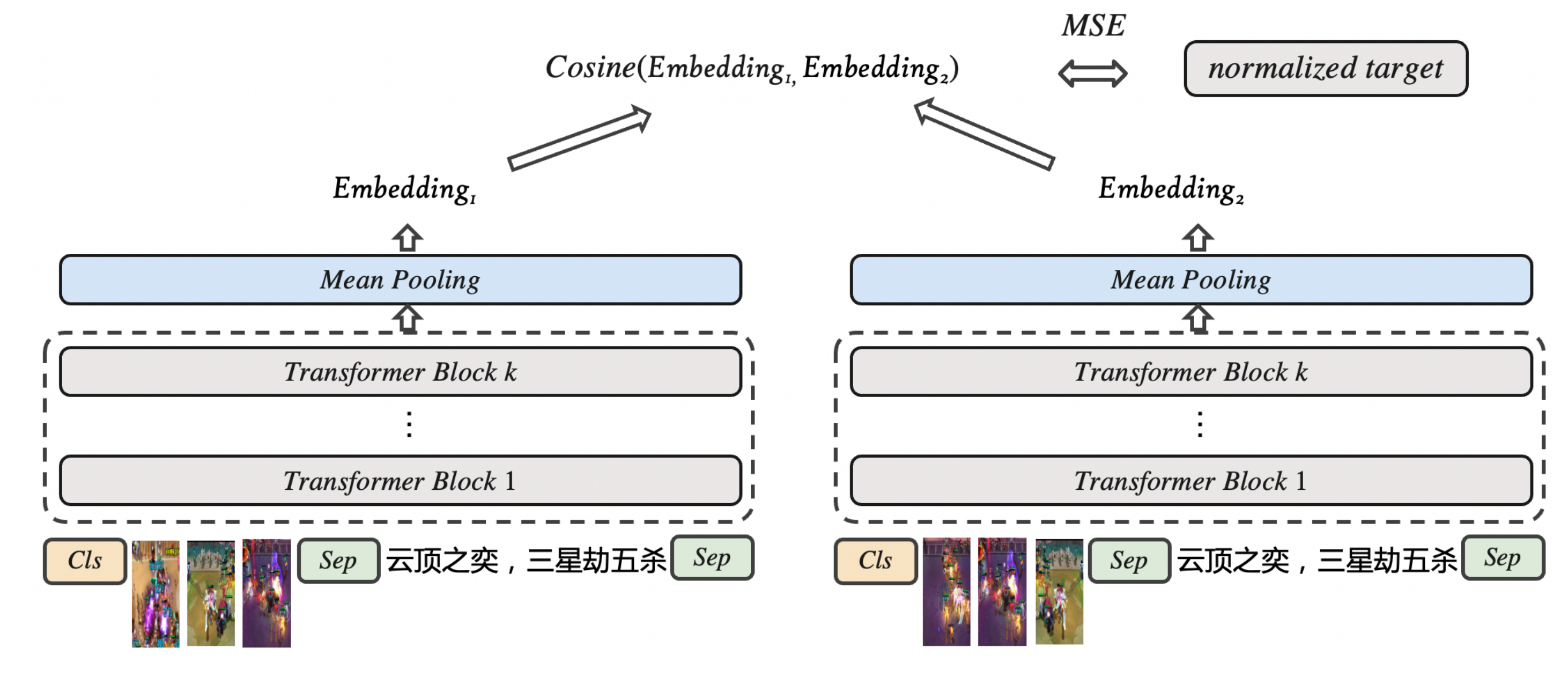}
\caption{Model structure for multi-modal similarity finetune.}
\label{fig1}
\vskip -0.01in
\end{figure*}

\renewcommand\arraystretch{}
\begin{table*}[!h]
\centering

\scalebox{1}{
\begin{tabular}{c | c |  c | c | c}
\toprule

No & Pretrain Data * Epoch & Pretrain Task \& Stucture & Valid Spearman & Valid MSE\\
\midrule
1 & (point + pair) * 40 & VTC + MLM + MFM & 0.8858 ± 0.0009 & 0.0288 ± 0.0005\\
2 & (point + pair) * 40 & VTC (mean\_pooling) + MLM + MFM & 0.8855 ± 0.0026 & 0.0291 ± 0.0008 \\
3 & (point + pair) * 40 & VTC (mean\_pooling) + MLM & 0.8828 ± 0.0020 & 0.0296 ± 0.0009\\
4 & (point + pair + test) * 32 & VTC + MLM + MFM & 0.8847 ± 0.0019 & 0.0290 ± 0.0009\\
5 & (point + pair) * 40 & VTC + MLM ($p_{mask}$=0.25) + MFM ($p_{mask}$=0.25) & 0.8859 ± 0.0025 & 0.0288 ± 0.0009\\
6 & (point + pair) * 40 -> pair * 5 & VTC + MLM + MFM & 0.8840 ± 0.0014 & 0.0293 ± 0.0009\\
\bottomrule
\end{tabular}}
\caption{Overview of model ensemble}
\label{table5}
\end{table*}
\vspace{-1cm}

\subsection{Finetune}
\subsubsection{Finetune tasks}
We add a pooling layer to extract video multi-modal embedding and a fc layer to reduce the feature dimension to 256. 
Video pair embedding is extracted separately and calculated cosine similarity as video similarity.
Finetune loss is the mean squared error (MSE) between video similarity and label.
The model structure for finetuning can be seen in figure 2.

\subsubsection{Rank normalization of targets}
Evaluation metrics is Spearman, which is calculated based on rank. 
We first rank the target, normalized to [0, 1] and then use them to compute loss.
Rank normalization improves our test Spearman's score about 0.006.

\subsubsection{Finetune details}
We finetune 10 epochs with 32 batch size. The learning rate for Bert backbone is 1e-5, and the learning rate of finetuning header is  5e-5. The learning scheduler is warm up with cosine decay, warm up ratio is 0.06.
   
\subsection{Ensemble}
\paragraph{Ensemble methods}
For blending our models, We first normalize the model's output embedding by using the l2 regularity, then concatenate them and use the singular value decomposition (SVD) method to reduce the embedding dimension to 256.
SVD-256 embedding Spearman is 0.001 lower than concated embedding.

\paragraph{Ensemble results}
Our final submission is an ensemble of 6 models. 
Each model is similar to the method described above, table 4 gives an overview of each ensemble model.
In the table, VTC with mean pooling means we change task header from $[CLS]$ + fc to mean\_pooling + fc.
$p_{mask}$ means we change mask prob in MLM and MFM from 0.15 to 0.25.

Our best single model score is 0.836 in testset. Ensemble 2 models achieve 0.845, 3 models achieve 0.849 and 5 models achieve 0.852.
            
\subsection{Validation Strategy}
We have 5 validation sets for the video similarity task. 
Each fold train - val  is split by video ids ensure that train videos and val videos are not overlapped.  

\begin{itemize}
\item $i_{th}$ fold train set : video pairs $(vid1, vid2)$ where $ vid_{1} \,\%\, 5\neq i$ and $ vid_{2} \,\%\, 5 \neq i $
\item $i_{th}$ fold valid set : video pairs $(vid1, vid2)$ where $ vid_{1} \,\%\, 5 = i$ and $ vid_{2} \,\%\, 5 = i $
\end{itemize}

The validation strategy is quite stable in the competition and we report downstream performance mean and std across these five validation sets.  


\section{Conclusion}
In this paper, we presented our winning solution to the 2021 AIAC competition track 1. 
The solution uses the multi-modal transformer models to extract video embedding, the model pretrain on VTC, MLM, MFM tasks, and finetune on video similarity task. Pretrain plays an important role in our solution.
After ensembling several models with different pretrain strategies, we reached a score of 0.852 on the leaderboard and take the first place.



\end{document}